\documentclass{ceurart}

\usepackage{booktabs}
\usepackage{multirow}
\usepackage{array}
\usepackage{listings}
\usepackage{xcolor}
\usepackage{amsmath}
\usepackage{amssymb}

\newcommand{\Fone}{F\textsubscript{1}}
\newcommand{\pp}{\,\textnormal{pp}}

\begin{document}

\copyrightyear{2026}
\copyrightclause{Copyright for this paper by its authors.
  Use permitted under Creative Commons License Attribution 4.0
  International (CC BY 4.0).}

\conference{CLEF 2026 Working Notes, 21--24 September 2026, Jena, Germany}

\title{Shrome at Touch\'e: Soft-Vote Ensembling and Counter-Causal Augmentation for Causality Extraction}

\subtitle{Touch\'e at CLEF 2026}

\author[1]{Roham Zendehdel Nobari}[email=roham.zendehdelnobari@uzh.ch]
\cormark[1]
\fnmark[1]
\author[1]{Shayan Sooratgar}[email=shayan.sooratgar@uzh.ch]
\fnmark[1]
\address[1]{Department of Computational Linguistics, University of Zurich, Switzerland}
\cortext[1]{Corresponding author.}
\fntext[1]{These authors contributed equally and share first authorship.}

\begin{abstract}
Touch\'e 2026 extends causality extraction to counter-causal
claims: news sentences whose surface form appears causal but
whose meaning denies the causation, as in \emph{``It is falsely
believed that X caused Y.''} A system that relies on surface
cues such as \emph{caused} or \emph{led to} will accept such a
sentence as causal and give it the wrong polarity. On the
Countercausal News Corpus (CCNC)~\cite{hagen:2025}, the task has
three subtasks: deciding whether a sentence is causal
(detection), locating its cause and effect spans (extraction),
and labeling its polarity as procausal, counter-causal, or
uncausal. We build one model per subtask. Detection is a fine-tuned
classifier with a single cross-task rule that uses the extracted
spans to remove false positives. For extraction, we ensemble
three RoBERTa-large~\cite{liu:2019roberta} BILOU+CRF taggers by
averaging their token-level scores before decoding, rather than
voting on the spans each tagger produces. For polarity, where
labeled counter-causal examples are scarcest, we add training
sentences generated by a large language model prompted with nine
patterns of counter-causal expression adapted from Hagen et
al.~\cite{hagen:2025}, keeping only those that pass automatic
structural checks. On the held-out CCNC test set, the system
reaches \Fone{} 0.869 on detection and macro-\Fone{} 0.817 on
polarity, and in the organizers' final causal-only evaluation of
extraction it scores granularity-adjusted \Fone{} 0.728, the
highest extraction score among all submissions including the
organizers' baseline. The development split is used only for
component selection and the ablations reported in the paper.
\end{abstract}


\maketitle

\section{Introduction}
\label{sec:intro}

Causal claims pervade news text, with reports routinely
asserting that one event brought about another. Automatically
extracting these claims supports downstream causal reasoning. The
difficulty is that many causal statements in public discourse
are disputed, outdated, or explicitly denied. A claim such as
\emph{``sugar causes hyperactivity''} is widely repeated yet
contradicted by the evidence, and a news sentence may invoke a
cause-effect pair precisely in order to reject it. Research on
causality extraction has largely neglected such
\emph{counter-causal} statements~\cite{hagen:2025}, even though
telling an asserted cause apart from a denied one is essential
to representing causal knowledge faithfully.

The Touch\'e 2026 lab on Argumentation
Systems~\cite{kiesel:2026c,kiesel:2026d} targets this gap with
its Task~2, Causality Extraction, built on the Countercausal
News Corpus (CCNC)~\cite{hagen:2025}. Relative to prior shared tasks
on the Causal News Corpus~\cite{tan:2022cnc,tan:2022case}, the
central change is the presence of counter-causal sentences whose
surface form contains an apparent cause-effect pair but whose
discourse \emph{denies} the causation (\emph{``It is falsely
believed that X caused Y.''}, \emph{``It is not that X caused
Y.''}). A model that detects causality from surface
lexical cues alone -- the verbs \emph{caused}, \emph{led to},
\emph{resulted in} -- will accept such sentences as causal and
assign the wrong polarity. The task decomposes into three
subtasks. \textbf{Detection} is binary classification of whether
the sentence is causal, \textbf{extraction} locates the cause
and effect spans, and \textbf{polarity} labels a given
sentence-and-pair as procausal, counter-causal, or uncausal.

We use \emph{procausal} for what Hagen et
al.~\cite{hagen:2025} call the \emph{causal} polarity class, to
avoid the overload with the detection subtask's binary
\emph{causal} label. We treat the three subtasks as separate
models with one cross-task interaction in which
extraction-stage spans gate the detection-stage positive output.
Detection itself is a fine-tuned BERT classifier with that
single gating rule, so our two methodological contributions fall
in extraction and polarity; the third reports the full system.
Our contributions, organized by subtask, are as follows.
\begin{enumerate}
\setlength{\itemsep}{0pt}
\item \textbf{Extraction (emission-level soft-vote
  ensembling).} A single
  RoBERTa-large~\cite{liu:2019roberta} model with a three-layer
  BILOU+CRF~\cite{schrader:2023} tagger, trained on the CCNC
  extraction split plus the external CNCv2/RECESS
  corpus~\cite{tan:2023recess} with EDA
  augmentation~\cite{wei:2019eda}, reaches 0.5742 exact-match
  \Fone{} on the CCNC development split before any ensembling;
  this base supplies most of our extraction performance. Our
  contribution is to combine three such seeds by averaging
  per-token CRF emissions and transition matrices before a
  per-layer Viterbi decode, rather than voting over
  already-decoded spans. On a matched three-seed panel this
  emission-level soft-vote beats span-level hard voting by
  $+1.07$\pp{} exact-match \Fone{}, with a parallel gain on the
  more lenient overlap metric, and the production panel reaches
  0.5933 (Section~\ref{sec:st2-ensemble}). Emission averaging is a
  standard CRF-ensembling pattern; the contribution is its
  matched-panel measurement on the Causal News Corpus family,
  where prior CASE submissions did not report this comparison.
\item \textbf{Polarity (taxonomy-conditioned counter-causal
  augmentation).} We generate counter-causal
  training data with a GPT-class large language model
  conditioned on nine patterns of counter-causal expression
  adapted from Hagen et al.'s Table~4 inventory and validate
  every candidate with mechanical structural checks. The structural-validator
  acceptance rate on the production generation pass is 93.2\%, a
  syntactic pass rate that does not certify semantic
  correctness. This augmentation is the main lever for polarity.
  At a fixed seed, two augmentation passes raise dev macro
  \Fone{} from 0.766 (no augmentation) to 0.890, with the
  per-pass breakdown reported in Section~\ref{sec:st3-aug}.
\item \textbf{Full system (all three subtasks).} Integrating the
  per-subtask models into a single pipeline, the system reaches
  binary \Fone{} 0.869 on detection and macro \Fone{} 0.817 on
  polarity on the held-out test set, and ranks first on
  extraction in the organizers' final causal-only evaluation
  with granularity-adjusted \Fone{} 0.728
  (Section~\ref{sec:test} reports the TIRA scores, development
  results, and the dev-to-test gap).
\end{enumerate}

\section{Related Work}
\label{sec:related}

\paragraph{Causality extraction and the Causal News Corpus.}
Recent work on causality extraction from news centers on the Causal News Corpus. CNCv1~\cite{tan:2022cnc} annotates English news sentences from Indian outlets for binary causal classification; its successor CNCv2, also released as RECESS~\cite{tan:2023recess}, adds cause/effect/signal span annotation. The two underpinned the CASE~2022~\cite{tan:2022case} and CASE~2023~\cite{tan:2023case} Event Causality Identification shared tasks. These resources treat a sentence as either asserting a causal relation or not, with no category for sentences that deny one. Hagen et al.~\cite{hagen:2025} address this gap with the Countercausal News Corpus, which extends CNCv2 by manually rewriting roughly half of its causal sentences into counter-causal form, producing a three-way procausal/counter-causal/uncausal corpus (1{,}028 / 952 / 1{,}435 sentences). They also provide RoBERTa baselines on its validation split (macro-averaged: detection \Fone{} 0.874, BIO-tag \Fone{} 0.440, ternary identification \Fone{} 0.921), computed under metric and split conventions that differ from those of the shared task and so are not directly comparable to results reported here.

\paragraph{Span extraction with encoder--CRF taggers.}
Pairing a neural encoder with a linear-chain CRF~\cite{lafferty:2001crf} over the BILOU tagging scheme~\cite{ratinov:2009bilou} has been the standard recipe for span extraction since the BiLSTM-CRF tagger of Lample et al.~\cite{lample:2016bilstmcrf}, later restated with transformer encoders in place of the BiLSTM. For causal spans specifically, the BoschAI submission to CASE~2023~\cite{schrader:2023} ranked first on Subtask~2 using a multi-layer BILOU+CRF tagger over a shared transformer encoder, stacking simultaneous cause-effect-signal relations through pipe-concatenated labels $\langle t_1 | t_2 | t_3 \rangle$ that admit up to three relations per sentence.

\paragraph{Data augmentation for low-resource labeling.}
Easy Data Augmentation (EDA)~\cite{wei:2019eda} enlarges training sets through synonym replacement and random word operations, but its sentence-level edits disturb token-level span boundaries; Dai and Adel~\cite{dai:2020eda4ner} instead design label-aware token-level operations for sequence labeling, replacing tokens or whole mentions in a way that keeps the tag sequence consistent. A separate line synthesizes new labeled examples with large language models, e.g., Schick and Sch{\"u}tze~\cite{schick:2021generating}, which is attractive for enlarging rare classes that have little annotated data.

\paragraph{NLI as a zero-shot classifier and verifier.}
Yin et al.~\cite{yin:2019zeroshot} showed that off-the-shelf natural-language-inference models can act as zero-shot text classifiers by phrasing each candidate label as an entailment hypothesis, and the same paradigm has since been used to verify generated or extracted text. Strong multi-corpus checkpoints, such as the DeBERTa-v3~\cite{he:2023debertav3} model of Laurer et al.~\cite{laurer:2023deberta} trained on MNLI~\cite{williams:2018mnli} and several further NLI datasets, make this usable without task-specific training.

\paragraph{Task-adaptive pre-training.}
Gururangan et al.~\cite{gururangan:2020dapt} show that continuing masked-language-model pre-training on in-domain or task text before fine-tuning improves downstream accuracy, a low-cost adaptation when the labeled target corpus is small.

\section{Dataset}
\label{sec:data}

CCNC~\cite{hagen:2025} extends CNCv2~\cite{tan:2023recess} by manually rewriting roughly half of CNCv2's causal sentences into counter-causal form. We train on the organizer-provided splits (Table~\ref{tab:datastats}); the extraction and polarity models additionally use in-domain augmentation, described in Sections~\ref{sec:st2} and~\ref{sec:st3}.

\begin{table}[htbp]
  \caption{CCNC counts on the splits used. Detection and extraction share sentence-level data; polarity has more rows because each candidate cause/effect pair within a sentence is annotated separately.}
  \label{tab:datastats}
  \centering
  \begin{tabular}{lrrr}
    \toprule
    Subtask & Train & Dev & Gold-span or class units \\
    \midrule
    Detection (binary)            & 3{,}075 & 340 & 1 label per sentence \\
    Extraction                    & 3{,}075 & 340 & 473 gold spans on dev \\
    Polarity (3-class)            & 4{,}603 & 502 & 2{,}412/1{,}305/886 train; 264/120/118 dev \\
    \bottomrule
  \end{tabular}
\end{table}

The polarity training set is class-skewed, and the smallest class, counter-causal, carries explicit negation cues (\emph{not}, \emph{n't}, \emph{never}, \emph{falsely}, \emph{contrary}, \emph{denied}) far more often than the other two (Table~\ref{tab:polaritycomp}). This roughly four-fold cue gap between procausal and counter-causal motivates the negation-scope feature discussed in Section~\ref{sec:discussion}.

\begin{table}[htbp]
  \caption{Polarity training composition. The counter-causal class is the smallest yet shows explicit negation cues most often (cue prevalence measured on dev). The nine-pattern LLM augmentation of Section~\ref{sec:st3} adds 2{,}433 rows, raising it from 886 to 3{,}319.}
  \label{tab:polaritycomp}
  \centering
  \begin{tabular}{lrrr}
    \toprule
    Class & Train rows & Neg.\ cues (dev) & After aug.\ \\
    \midrule
    uncausal       & 2{,}412 & 29.5\% & 2{,}412 \\
    procausal      & 1{,}305 & 15.8\% & 1{,}305 \\
    counter-causal &    886 & 67.8\% & 3{,}319 \\
    \midrule
    Total          & 4{,}603 & ---    & 7{,}036 \\
    \bottomrule
  \end{tabular}
\end{table}

\section{System Overview}
\label{sec:overview}

\begin{figure}[htbp]
  \centering
  \includegraphics[width=0.85\linewidth]{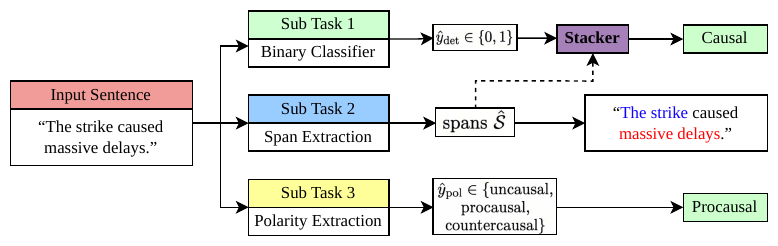}
  \caption{System overview on an illustrative input sentence. The detection, extraction, and polarity models are trained independently; the only cross-task interaction is the dashed arrow, through which the predicted spans $\hat{\mathcal{S}}$ gate the detection output via the one-rule stacker (causal iff $\hat y_{\text{det}}=1$ and $|\hat{\mathcal{S}}|\geq 1$).}
  \label{fig:pipeline}
\end{figure}

Three subtask models run independently on the same input sentence (Figure~\ref{fig:pipeline}): a BERT-base binary detector, a three-seed emission-level soft-vote ensemble of three-layer BILOU+CRF taggers on RoBERTa-large for extraction, and a four-seed mean-softmax ensemble of BERT-base classifiers with PURE typed entity markers around cause and effect spans for polarity. A single-rule cross-task stacker overrides detection to \emph{uncausal} whenever extraction produces no spans, correcting about twice as many dev false positives as the new false negatives it introduces (exact flip counts in Section~\ref{sec:st1-stacker}), without retraining either model.

\section{Detection}
\label{sec:st1}

\subsection{Architecture, training, and the cross-task stacker}
\label{sec:st1-stacker}

\begin{figure}[htbp]
  \centering
  \includegraphics[width=0.9\linewidth]{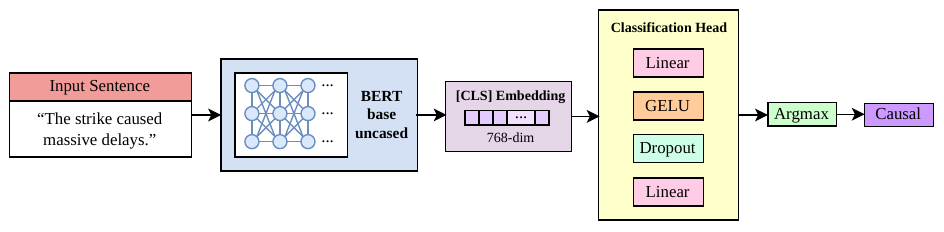}
  \caption{Detection model. The input sentence is encoded by
  BERT-base; the \texttt{[CLS]} representation passes through the
  stock sequence-classification head, and an argmax over the two
  logits yields the binary causal/uncausal label.}
  \label{fig:st1-arch}
\end{figure}

We fine-tune \texttt{bert-base-uncased}~\cite{devlin:2019bert} with
the default sequence-classification head over the pooler output
(a tanh layer on the \texttt{[CLS]} hidden state), shown in
Figure~\ref{fig:st1-arch}. Inputs are raw
sentences truncated at 128 WordPiece tokens.

 Training is a standard fine-tune at a single seed
(hyperparameters in Appendix~\ref{app:hyper}). No early stopping
is configured: the model trains a fixed five epochs and the
best-validation-loss checkpoint is restored at the end. We did not invest heavily in detection: a TF-IDF plus
logistic-regression baseline already reaches 0.793 binary
\Fone{} on dev, leaving headroom that BERT-base closes
(Table~\ref{tab:st1results}).

The cross-task stacker is a one-rule override that flips the
detector's positive output to \emph{uncausal} when the extraction
stage predicts no spans: if no cause or effect span is located,
no causal relation can be reported. Formally, we predict
\emph{causal} iff $\hat y_{\text{det}}=1$ \emph{and}
$|\hat S|\geq 1$, and \emph{uncausal} otherwise. This rule was
selected by grid search over ten candidates on the dev split:
one detector-only rule plus disjunctive (OR), flip, and
conjunctive (AND) variants parameterized by the minimum span
count. Over the standalone detector (Table~\ref{tab:st1results}), the
conjunctive rule with span threshold one maximizes dev
causal-class \Fone{} at 0.8946, a 2.6\pp{} gain.

The rule flips 46 dev sentences from causal to uncausal: 31
flips correct existing false positives and 15 introduce new
false negatives. The calibration log accordingly records
causal-class recall over the 207 gold causal sentences dropping
from 0.9130 standalone to 0.8406 stacked, a recall loss that the
precision gain outweighs on this split. The rule is only safe
while extraction recall on causal sentences stays high enough
that the span-presence gate does not start removing true
positives. With ten candidates evaluated on a 340-sentence dev
split, the gain partly reflects selection over the rule space. We did not submit the
unstacked detector to the TIRA test split, so we cannot
quantify how much of the dev gain transferred; the stacked
test causal-class \Fone{} of 0.869
(Section~\ref{sec:test}) is consistent with either partial
transfer plus underlying detector drift, or full transfer on
a harder underlying split.

\subsection{Baselines and dev results}

Table~\ref{tab:st1results} summarizes the detection dev results.
The simple TF-IDF baseline outperforms the BiLSTM as well as both
off-the-shelf causal systems (the UniCausal sequence baseline and
zero-shot BART-MNLI), fine-tuned BERT-base clears every baseline
by a wide margin, and the span-gated stacker of
Section~\ref{sec:st1-stacker} adds the final increment and is the
production configuration.

\begin{table}[htbp]
  \caption{Detection dev results (binary \Fone{}). The production
  system is the stacked variant in the last row. UniCausal
  baseline is the sequence-classification head of Tan et
  al.~\cite{tan:2023unicausal}; BART-MNLI is the zero-shot
  pipeline over \texttt{facebook/bart-large-mnli}.}
  \label{tab:st1results}
  \centering
  \begin{tabular}{lr}
    \toprule
    Configuration & Binary \Fone{} (causal class) \\
    \midrule
    Random                                   & 0.490 \\
    TF-IDF + logistic regression             & 0.793 \\
    BiLSTM                                   & 0.731 \\
    UniCausal sequence baseline              & 0.553 \\
    Zero-shot BART-MNLI                      & 0.642 \\
    BERT-base (seed 42, standalone)          & 0.8690 \\
    \midrule
    \textbf{Stacker: BERT-base $\wedge$ (extraction has spans)} & \textbf{0.8946} \\
    \bottomrule
  \end{tabular}
\end{table}

\section{Extraction}
\label{sec:st2}

\subsection{Per-model architecture: three-layer BILOU+CRF}
\label{sec:st2-arch}

\begin{figure}[htbp]
  \centering
  \includegraphics[width=0.95\linewidth]{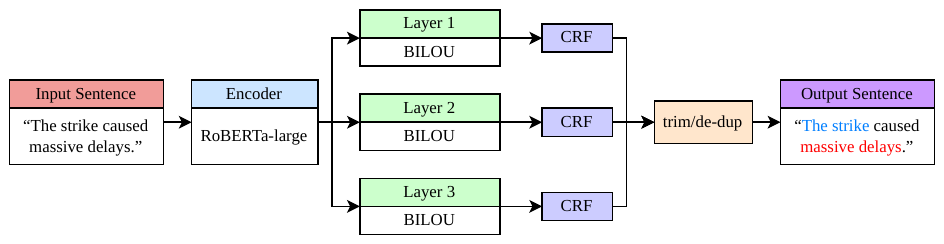}
  \caption{One extraction seed. A shared RoBERTa-large encoder
  feeds three parallel BILOU layers, each decoded by its own
  linear-chain CRF; the decoded spans are unioned across layers,
  de-duplicated, and trimmed to produce the output spans.}
  \label{fig:st2-model}
\end{figure}

Each seed adapts the multi-layer BILOU+CRF idea of
BoschAI~\cite{schrader:2023} into a parallel-head form
(Figure~\ref{fig:st2-model}): three
independent linear BILOU heads on a shared RoBERTa-large
encoder~\cite{liu:2019roberta}, each producing five-way per-token
logits over $\{O, \text{B-ENT}, \text{I-ENT}, \text{L-ENT}, \text{U-ENT}\}$ -- Outside, Begin, Inside, Last, and Unit (single-token span) -- for a
single ENTITY type (CCNC's extraction-stage annotation does not
distinguish cause from effect, so one label suffices) and decoded
by its own linear-chain CRF~\cite{lafferty:2001crf}; in contrast
to BoschAI's
pipe-concatenated composite labels (Section~\ref{sec:related}),
the three layers do not share output space or transition
matrix. Gold spans within a sentence are
greedily colored onto the three layers by an interval-graph
algorithm; no CCNC training sentence requires more than three
layers, so none are dropped. The loss averages the per-layer
negative CRF log-likelihood, and for the TAPT seed adds a
uniform label-smoothing auxiliary cross-entropy: a
simplification of the targeted boundary-smoothing scheme of Zhu
and Li~\cite{zhu:2022bs} that retains the regularization intent
without boundary-localized reweighting. Two training choices
matter beyond defaults: the CRF parameters sit in a separate
optimizer group at a roughly six-fold higher learning rate than
the encoder, and early stopping monitors overlap \Fone{} rather
than validation loss. All hyperparameter values are listed in
Appendix~\ref{app:hyper}.

\paragraph{Training data.}
CCNC alone is small enough that the tagger overfits it: a
RoBERTa-large model trained on the $3{,}075$-row extraction
split without augmentation, with validation-loss early stopping,
reaches only 0.40 exact \Fone{} on dev (training loss collapses
while recall stalls). Adding external in-domain data and EDA
augmentation, and switching the early-stopping monitor from
validation loss to overlap \Fone{}, together lift the
best single seed to 0.5742 (Table~\ref{tab:st2-ensemble}); we
did not run a one-factor-at-a-time ablation, so we report these
as a joint effect.

The ensemble averages three independently trained seeds, where a
\emph{seed} is one complete fine-tuning run
(Section~\ref{sec:st2-ensemble}): two follow a \emph{base recipe}
and one a \emph{TAPT recipe} (seeds listed in
Appendix~\ref{app:hyper}). Both
recipes add CNCv2 (RECESS)~\cite{tan:2023recess}, converted to the
CCNC schema with signal tokens stripped (keeping the signal
annotations costs about 6 points of overlap \Fone{}, as the model
then spends capacity on a span class absent from CCNC), and apply
EDA~\cite{wei:2019eda} (synonym replacement and random
insertion; one augmented copy per sentence with fewer than three
spans) plus a double oversample of sentences with three or more
spans. The TAPT recipe differs in
three ways: it (i)~warm-starts from a \emph{task-adaptively
pre-trained} encoder~\cite{gururangan:2020dapt}, that is, one
produced by continued masked-language modeling on the unique
CCNC and CNCv2 sentences, driving the masked loss from $1.67$
to $0.36$ (schedule and corpus size in
Appendix~\ref{app:hyper}), rather than off-the-shelf
RoBERTa-large; (ii)~concatenates the
UniCausal~\cite{tan:2023unicausal} grouped rows; and
(iii)~adds the boundary-smoothing loss described above.
Table~\ref{tab:extractionpool} gives the resulting pool sizes.

\begin{table}[htbp]
  \caption{Extraction training-pool composition by recipe. Source
  rows are expanded by EDA augmentation and multi-span
  oversampling; the two base seeds differ slightly (seed 123:
  $11{,}489$; seed 2026: $11{,}492$ rows) and the TAPT seed
  reaches $13{,}432$.}
  \label{tab:extractionpool}
  \centering
  \begin{tabular}{lrr}
    \toprule
    Source & Base recipe & TAPT recipe \\
    \midrule
    CCNC extraction train            & 3{,}075 & 3{,}075 \\
    CNCv2/RECESS (signals stripped)  & 3{,}075 & 3{,}075 \\
    UniCausal (signals stripped)     & ---     & 1{,}049 \\
    \midrule
    Source rows                      & 6{,}150 & 7{,}199 \\
    After EDA + oversampling         & $\approx$11{,}500 & $\approx$13{,}400 \\
    \bottomrule
  \end{tabular}
\end{table}

\paragraph{Decoding.}
At inference we collect per-layer BILOU logits, Viterbi-decode
each layer through its own CRF, map tag sequences to
character-level spans via the RoBERTa offset map, and union
spans across the three layers, deduplicating exact matches and
trimming whitespace and the $15$ punctuation characters
\verb|"|, \verb|'|, \verb|`|, \verb|,|, \verb|.|, \verb|;|,
\verb|:|, \verb|!|, \verb|?|, \verb|(|, \verb|)|, \verb|[|,
\verb|]|, \verb|{|, \verb|}| from span endpoints.

\subsection{Emission-level soft-vote across three seeds}
\label{sec:st2-ensemble}

\begin{figure}[htbp]
  \centering
  \includegraphics[width=\linewidth]{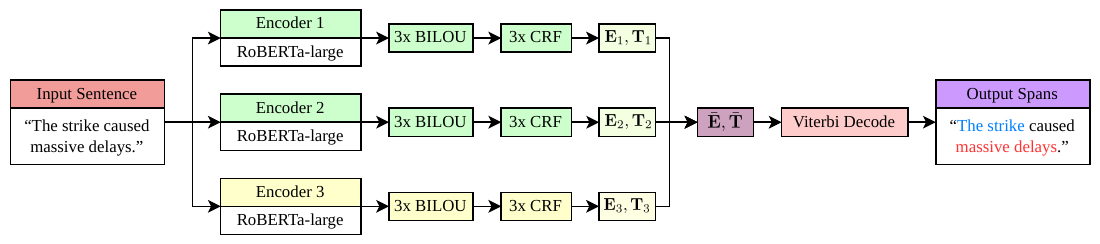}
  \caption{Emission-level soft-vote. Each of the three seeds
  produces per-layer emissions and transition scores
  $(\mathbf{E}_k, \mathbf{T}_k)$; these are averaged into
  $(\bar{\mathbf{E}}, \bar{\mathbf{T}})$ \emph{before} a single
  per-layer Viterbi decode, so near-miss agreement between seeds
  survives to the decoding step instead of being discarded by
  span-level voting. The two green encoders are the base-recipe
  seeds; the yellow encoder is the seed warm-started from the
  task-adaptively pre-trained (TAPT) checkpoint
  (Section~\ref{sec:st2-arch}).}
  \label{fig:st2-softvote}
\end{figure}

The final ensemble averages CRF emissions and transition matrices
across three seeds before a per-layer Viterbi decode through each
averaged CRF, rather than voting over predicted spans
(Figure~\ref{fig:st2-softvote}). A single seed of the
architecture above reaches dev exact \Fone{} of 0.5742 and
overlap \Fone{} of 0.7692 (seed 123); hard span-level voting
across three or four seeds adds only a marginal exact-\Fone{}
gain (Table~\ref{tab:st2-progression}). Let $N$ denote the number of subword tokens in the sentence and
identify the BILOU tag set $\{O,B,I,L,U\}$ with $\{1,\ldots,5\}$.
Given $K$ models each producing $L$ layers of emissions
$\mathbf{E}_k^{(\ell)}\in\mathbb{R}^{N\times 5}$, transition
matrices $\mathbf{T}_k^{(\ell)}\in\mathbb{R}^{5\times 5}$, and
start and end vectors
$\mathbf{s}_k^{(\ell)},\mathbf{e}_k^{(\ell)}\in\mathbb{R}^5$,
define the cross-seed mean
$\bar{\mathbf{X}}^{(\ell)}=K^{-1}\sum_{k=1}^{K}\mathbf{X}_k^{(\ell)}$
for each tensor. With $K=L=3$ in our system, we Viterbi-decode
\[
\arg\max_{y_{1:N}\in\{1,\ldots,5\}^{N}}\;
\bar{\mathbf{s}}^{(\ell)}_{y_1}
+\sum_{n=1}^{N} \bar{\mathbf{E}}^{(\ell)}_{n,y_n}
+\sum_{n=1}^{N-1} \bar{\mathbf{T}}^{(\ell)}_{y_n,y_{n+1}}
+\bar{\mathbf{e}}^{(\ell)}_{y_N}
\]
separately for each layer $\ell\in\{1,\ldots,L\}$, with span
union across the $L$ decoded sequences and the trimming of
Section~\ref{sec:st2-arch}.

Emission-level voting recovers near-miss agreement that
span-level voting discards: when models A, B, and C assign
START probabilities of 0.80, 0.60, and 0.70 at position $n$,
the averaged decode sees 0.70 even if no individual decoder
emitted a START there. On a matched three-seed panel, swapping
span-level hard-vote for emission-level soft-vote adds 1.07\pp{}
exact and a comparable overlap gain
(Table~\ref{tab:st2-progression}). Averaging CRF tensors before joint Viterbi is a
standard CRF-ensembling pattern; the BoschAI
submission~\cite{schrader:2023} fine-tuned a single multi-layer
BILOU+CRF model without ensembling, and to our knowledge
emission-level soft-vote across seeds on multi-layer BILOU+CRF
panels has not been reported for the Causal News Corpus family.

\paragraph{The production panel.}
After a sweep over candidate triples we select seeds 123 and
2026 on the base recipe and seed 42 on the TAPT recipe with the
boundary-smoothing auxiliary loss, the triple that maximizes
the mean of exact and overlap \Fone{} on dev
(Table~\ref{tab:st2-ensemble}). This panel differs from the
matched comparison above because one seed carries the TAPT
warm-start, so its endpoint cannot be read directly from the
hard-versus-soft delta. The TAPT recipe helped only seed 42
(+1.17\pp{} exact) and hurt the other three seeds we tested by
1.6\pp{} to 3.0\pp{}, averaging $-1.38\pp{}$ across all four
paired seeds. We retain the TAPT seed for emission diversity
through a different encoder warm-start, not as a method-level
endorsement of TAPT (Section~\ref{sec:discussion}).

\begin{table}[htbp]
  \caption{Per-seed and ensemble dev \Fone{} for the production
  extraction panel.}
  \label{tab:st2-ensemble}
  \centering
  \begin{tabular}{lcccc}
    \toprule
    Seed & Encoder warm-start & Recipe & Exact \Fone{} & Overlap \Fone{} \\
    \midrule
    123 & RoBERTa-large       & base recipe                & 0.5742 & 0.7692 \\
    2026 & RoBERTa-large      & base recipe                & 0.5505 & 0.7569 \\
    42 & RoBERTa-large + TAPT & +UniCausal, boundary-smoothing aux. & 0.5260 & 0.7584 \\
    \midrule
    \multicolumn{3}{l}{Soft-vote of the three seeds above}    & \textbf{0.5933} & \textbf{0.7800} \\
    \bottomrule
  \end{tabular}
\end{table}

\subsection{NLI span verifier}
\label{sec:st2-nli}

As a post-hoc complement to the supervised pipeline, an
off-the-shelf entailment model independently accepts or rejects
each span produced by the soft-vote panel. For each span $s$ we
form the hypothesis
\verb|"This sentence describes a causal event involving '|$s$\verb|'."|
and feed (premise = original sentence, hypothesis) into a
multi-corpus DeBERTa-v3-large NLI
checkpoint~\cite{laurer:2023deberta} trained on MNLI plus four
additional NLI corpora, retaining spans with entailment
probability at least $\tau$. The threshold $\tau=0.1$ maximizes
exact \Fone{} on dev, removing only the three most clearly
spurious spans; thresholds $\tau\geq 0.2$ degrade \Fone{}
monotonically by removing borderline true spans. An earlier
pair-level template
\verb|"|$s_{\text{cause}}$ \verb|caused| $s_{\text{effect}}$\verb|"|
lost $5.8\pp{}$ \Fone{}: gold CCNC cause and effect
spans do not compose into well-formed directional hypotheses.
The pair \emph{``agitating ex-servicemen accusing the
government''}, for example, receives an entailment probability
of only 0.25 under that template.

Applied to the soft-vote panel, the verifier produces a
within-noise change on dev: it removes three predicted spans,
each a truncated fragment of a gold span, trading a marginal
exact-\Fone{} gain for a comparable overlap-\Fone{} loss
(Table~\ref{tab:st2-progression}). We discuss this within-noise
gain in Section~\ref{sec:discussion}.

\subsection{Final results}

Table~\ref{tab:st2-progression} tracks the dev progression:
emission-level soft-vote lifts the single-seed baseline by
$1.91\pp{}$ exact to the production panel's 0.5933, and the NLI
verifier adds a final step that sits below per-span granularity
on dev (Section~\ref{sec:discussion}). The Hagen et al.\ 0.440
row is BIO-tag macro \Fone{} on Hagen's Val split rather than
span exact-match on dev, so the $\Delta$ column is indicative
only; likewise our held-out test \Fone{} (Table~\ref{tab:test})
cannot be compared cleanly to Hagen's number because Hagen
releases no test split, so the apparent gap mixes splits and
metrics.

\begin{table}[htbp]
  \caption{Extraction dev progression. All numbers in this table
  except the first row are span-level exact-match \Fone{} on the
  CCNC dev split (overlap \Fone{} in parentheses).
  $^\dagger$The Hagen RoBERTa baseline is BIO-tag macro \Fone{}
  on Hagen's evaluation split (Hagen et al.\ Table~7; their
  Table~6a labels this as Val, with no separate test column),
  not span exact-match on dev; the two metrics and the two splits differ,
  so $\Delta$ is an indicative reference, not a directly
  comparable gain.}
  \label{tab:st2-progression}
  \centering
  \begin{tabular}{lrr}
    \toprule
    Stage & Exact \Fone{} (overlap \Fone{}) & $\Delta$ exact \\
    \midrule
    Hagen RoBERTa baseline~\cite{hagen:2025} (Val, BIO-tag) & 0.440 (n/a) & ref$^\dagger$ \\
    Single seed, base recipe (seed 123) & 0.5742 (0.7692) & $+13.4\pp$ \\
    Hard span-level vote, matched three-seed panel & 0.5781 (0.7634) & $+13.8\pp$ \\
    Emission soft-vote, same matched panel & 0.5888 (0.7730) & $+14.9\pp$ \\
    \textbf{Emission soft-vote, production panel with TAPT seed} & \textbf{0.5933} (0.7800) & $\mathbf{+15.3\pp}$ \\
    \quad + NLI span verifier ($\tau=0.1$, within-noise) & 0.5953 (0.7781) & $+15.5\pp$ \\
    \bottomrule
  \end{tabular}
\end{table}

\section{Polarity Classification}
\label{sec:st3}

\subsection{Architecture}

\begin{figure}[htbp]
  \centering
  \includegraphics[width=0.45\linewidth]{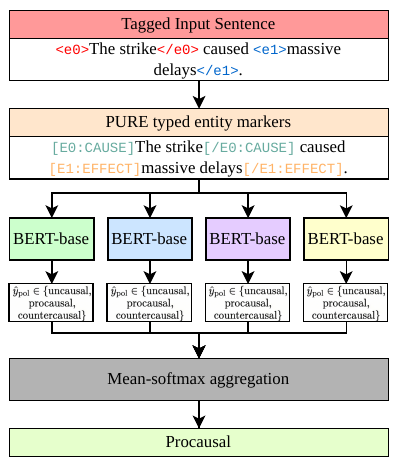}
  \caption{Polarity stack. The tagged input sentence is rewritten
  with PURE typed entity markers and classified by four
  independently seeded BERT-base models; their softmax outputs
  are averaged (mean-softmax) to produce the final three-way
  label.}
  \label{fig:st3-arch}
\end{figure}

We use \texttt{bert-base-uncased}~\cite{devlin:2019bert} with the default sequence-classification head (Figure~\ref{fig:st3-arch}). The one architectural change is to rewrite the bare entity tags as PURE typed entity markers~\cite{zhong:2021pure}: $\texttt{<e0>X</e0>}$ becomes $\texttt{[E0:CAUSE] X [/E0:CAUSE]}$ and $\texttt{<e1>Y</e1>}$ becomes $\texttt{[E1:EFFECT] Y [/E1:EFFECT]}$. Four special tokens are registered, the embedding table is resized, and the marker embeddings are learned from scratch during fine-tuning. Typed markers contribute a 0.8\pp{} macro \Fone{} gain on dev under the same recipe.

\subsection{The nine-pattern counter-causal augmentation}
\label{sec:st3-aug}

The main contribution for polarity is on the training-data side: we generate counter-causal training rows from a large language model conditioned on nine patterns of counter-causal expression adapted from Hagen et al.~\cite{hagen:2025} Table~4: \emph{direct\_negation, negated\_context, lack\_of\_counterfactuality, lack\_of\_effect, implicit\_lack\_of\_effect, negative\_causation, usual\_inverse\_effect, coincidence, temporal\_order}. Seven patterns instantiate Hagen Table~4 categories one-to-one ((h) direct negation, (i) negated context, (b) violation of counterfactuality, (c) lack of effect, (f) negative causation, (e) usual inverse effect, (a) violation of temporal order). The remaining two refine surface realizations Hagen lists as examples within existing categories rather than splitting them into separate categories: \emph{coincidence} corresponds to Hagen's (b) example expression ``A and B happened coincidentally'' and \emph{implicit\_lack\_of\_effect} corresponds to Hagen's (c) implicit example ``He is happy. He did not win the lottery.'' Hagen categories (d) inverse expected cause and (g) contradiction were not implemented; (d) is present in Hagen Table~6c's empirical distribution but its surface form overlaps (c) and (e), while (g) is absent from Table~6c. Hagen et al.\ explicitly assign \emph{negative\_causation} ($A$ \emph{prevents} $B$) to the \emph{uncausal} class rather than the counter-causal class (their Section~3.1, paragraph below Table~4: ``ontological asymmetry (f) \ldots\ indicates uncausality rather than countercausality''), and (f) does not appear in the empirical distribution of counter-causal expressions in CCNC (their Table~6c). We nevertheless retain (f) as a generation pattern because the resulting sentences share the negated-effect surface form that the polarity model must disambiguate from true counter-causal cases; this is a deliberate decision to inject hard surface-form confusables into the counter-causal class and is a known label-noise risk discussed in Section~\ref{sec:limitations}.

For each pattern, a one-shot prompt with a single fixed exemplar and an in-domain seed sentence from CCNC train produces candidates, which are then mechanically validated for byte-equal preservation of the \texttt{<e0>...</e0>} and \texttt{<e1>...</e1>} content and for the presence of exactly one \texttt{<e0>} pair and one \texttt{<e1>} pair. The structural-validator acceptance rate is 93.2\% (2{,}433 of 2{,}610 attempts on the production doubling pass; the smaller pilot pass was similar). Because the validators check only tag preservation and pair-count, this rate is an upper bound on usable yield rather than a semantic quality measure. A doubling augmentation pass grows the counter-causal training class from 886 to 3{,}319 rows. A seed-42 single-run dev ablation isolates the augmentation effect; each run is checkpointed at the epoch that scores best on a chosen validation metric. Augmentation raises dev macro \Fone{} from 0.766 (no augmentation, checkpoint selected by lowest validation loss) to 0.890 (two passes, checkpoint selected by highest validation macro \Fone{}, the criterion the deployed models use). Because these endpoints were checkpointed under different criteria, the controlled comparison is one pass against two under matched \Fone{} selection, where the second pass adds $+2.4$\pp{}. Each cell is a single fine-tuning run at seed 42, and the gain is of the same order as the per-seed noise band reported with the ensemble results below, so we treat it as direction-consistent rather than statistically certified. The deployed four-seed ensemble is built from the two-pass-augmented models (Table~\ref{tab:st3results}).

\subsection{Four-seed ensemble (with offline calibration)}

We train four BERT-base models with seeds $\{42, 123, 2026, 7\}$
under a shared recipe, early-stopping each on dev macro \Fone{}
(hyperparameters in Appendix~\ref{app:hyper}). For calibration
analysis, each model's softmax is rescaled by a single per-model
temperature $T$ fitted by LBFGS on the development set to
minimize negative log-likelihood, following Guo et
al.~\cite{guo:2017calibration}. Fitted temperatures fall in
$T\in[2.18, 2.39]$, and per-model calibration substantially
reduces both the negative log-likelihood and the expected
calibration error. Because the four models share architecture
and training recipe, temperature scaling shifts the ensemble
argmax only on a handful of borderline cases and does not
affect the ensemble dev macro \Fone{}
(Table~\ref{tab:st3results}). The deployed system
(Section~\ref{sec:test}) therefore
runs with $T=1.0$; the per-model temperatures are an offline
analysis on the development split.

\subsection{Results}

\begin{table}[htbp]
  \caption{Polarity dev results. Per-class \Fone{} on the right.
  For the production ensemble row, only the macro and
  counter-causal \Fone{} are reported: the stored uncausal and
  procausal per-class numbers are rounded to three decimals and
  do not reproduce the full-precision macro \Fone{} of 0.9029
  when averaged at that precision.}
  \label{tab:st3results}
  \centering
  \begin{tabular}{lrrrr}
    \toprule
    Configuration & Macro \Fone{} & uncausal \Fone{} & procausal \Fone{} & counter-causal \Fone{} \\
    \midrule
    Seed 42                          & 0.8897 & 0.9275 & 0.8678 & 0.8739 \\
    Seed 123                         & 0.8773 & 0.9091 & 0.8319 & 0.8908 \\
    Seed 2026                        & 0.8805 & 0.9246 & 0.8468 & 0.8703 \\
    Seed 7                           & 0.8867 & 0.9216 & 0.8571 & 0.8814 \\
    \midrule
    \textbf{4-seed mean-softmax (production)} & \textbf{0.9029} & --- & --- & \textbf{0.891} \\
    5-seed mean-softmax (adds a weaker seed) & 0.8953 & --- & --- & --- \\
    \bottomrule
  \end{tabular}
\end{table}

Table~\ref{tab:st3results} reports the dev results. The
four-seed ensemble macro \Fone{} of 0.9029 sits one to two
points above the best single seed and the four-seed mean
(per-seed scores in Table~\ref{tab:st3results}; four-seed mean
0.8836, sample standard deviation 0.0057, an estimate that is
itself uncertain at $n=4$). Without a held-out
distribution of four-seed ensembles, we report the gain as a
point improvement rather than a statistically certified one.
Adding a fifth seed with a weaker single-seed macro \Fone{} of
0.873 reduced the ensemble score (Table~\ref{tab:st3results}),
so we cap the ensemble at four members. This cap, the
dev-fit temperatures, and the dev-set seed pool are all
decisions made on the same evaluation split with no held-out
validation. For context, Hagen et al.~\cite{hagen:2025} Table~7
reports a RoBERTa baseline (size unspecified) for the
entity-pair Identification task at macro \Fone{} 0.921 on their
Val split (no separate test split is released; see their
Table~6a). Their task evaluates candidate entity pairs with
labels \{no-relationship, causal, countercausal\}, whereas our
polarity task evaluates dev pairs with labels
\{procausal, countercausal, uncausal\}, so the two numbers are
reference points rather than a head-to-head comparison. Our
four-seed BERT-base ensemble reaches 0.9029 on dev and 0.817 on
test. Two single-seed RoBERTa-large polarity pilots reached
macro \Fone{} 0.717 without augmentation and 0.726 with a
1{,}246-row pilot-era nine-pattern augmentation; the augmented pilot failed in a
wrong-direction-bias mode (high counter recall with low counter
precision) rather than classical overfitting, and we did not
pursue RoBERTa-large further. Closing this gap is the most
important open question for our polarity system.

\paragraph{Confusion matrix.}
The seed-42 confusion matrix on 502 dev sentences
(Table~\ref{tab:st3-confusion}) shows that the largest residual
error class is the 21 uncausal sentences predicted as either
procausal or countercausal. Mutual confusion between procausal
and countercausal is small, at five to six mistakes per
direction; the no-augmentation ablation in
Section~\ref{sec:st3-aug} shows that this gap widens without the
nine-pattern data.

\begin{table}[htbp]
  \caption{Polarity seed-42 confusion matrix on the dev split.
  Rows are gold; columns are predicted. Recall on the right.}
  \label{tab:st3-confusion}
  \centering
  \begin{tabular}{lrrrr}
    \toprule
    Gold $\downarrow$ / Predicted $\rightarrow$ & uncausal & procausal & counter-causal & recall \\
    \midrule
    uncausal (264)      & 243 & 12  & 9   & 0.920 \\
    procausal (120)     & 8   & 106 & 6   & 0.883 \\
    counter-causal (118) & 8   & 5   & 105 & 0.890 \\
    \bottomrule
  \end{tabular}
\end{table}

\section{Held-Out Test-Set Results}
\label{sec:test}

The Touch\'e organizers evaluated each submission on a held-out CCNC test split using the official TIRA~\cite{froebe:2023} evaluator. Table~\ref{tab:test} reports the test-set scores for our three subtask submissions alongside the matched development numbers from the per-subtask sections.

\begin{table}[htbp]
  \caption{Held-out CCNC test-set results, with matched dev numbers. Rows marked TIRA are from the official TIRA~\cite{froebe:2023} evaluator; the ``final, causal-only'' extraction rows are from the organizers' post-deadline re-evaluation of the extraction subtask on the causal test samples only, which raises scores without reordering submissions and will be reported in the lab overview~\cite{kiesel:2026d}. For extraction the TIRA evaluator reports exact-match \Fone{} together with a token-level IoU; the dev overlap \Fone{} is a different relaxation of exact match and is not directly comparable. $\Delta$ is test $-$ dev where metrics align. Row-internal arithmetic does not reconcile: the TIRA extraction \Fone{} (0.473) is a per-sentence \Fone{} averaged over the test set while the displayed precision (0.475) and recall (0.490) are corpus-level micro values, whose harmonic mean would be 0.482; the polarity macro \Fone{} (0.817) is the mean of per-class \Fone{} scores rather than the harmonic mean of macro precision (0.796) and macro recall (0.851), which would be 0.823. We report values as published.}
  \label{tab:test}
  \centering
  \small
  \begin{tabular}{llrrr}
    \toprule
    Subtask & Metric & Dev & Test & $\Delta$ \\
    \midrule
    Detection (binary)   & \Fone{} (causal class)     & 0.8946 & 0.869 & $-2.6\pp$ \\
    \quad                & precision                  & ---    & 0.848 & --- \\
    \quad                & recall                     & ---    & 0.890 & --- \\
    \quad                & accuracy                   & ---    & 0.832 & --- \\
    \midrule
    Extraction (TIRA)    & exact-match \Fone{}        & 0.5933 & 0.473 & $-12.0\pp$ \\
    \quad                & precision                  & ---    & 0.475 & --- \\
    \quad                & recall                     & ---    & 0.490 & --- \\
    \quad                & token IoU                  & ---    & 0.451 & --- \\
    \quad                & granularity                & ---    & 0.583 & --- \\
    \quad                & granularity-adjusted \Fone{} & ---  & 0.453 & --- \\
    Extraction (final, causal-only) & granularity-adjusted \Fone{} & --- & 0.728 & --- \\
    \quad                & precision                  & ---    & 0.764 & --- \\
    \quad                & recall                     & ---    & 0.787 & --- \\
    \midrule
    Polarity (3-class)   & macro \Fone{}              & 0.9029 & 0.817 & $-8.6\pp$ \\
    \quad                & macro precision            & ---    & 0.796 & --- \\
    \quad                & macro recall               & ---    & 0.851 & --- \\
    \bottomrule
  \end{tabular}
\end{table}

The dev-to-test gap orders as extraction (12.0\pp{} exact \Fone{}) > polarity (8.6\pp{} macro \Fone{}) > detection (2.6\pp{}), which is consistent with the dev-set overfitting disclosed in Section~\ref{sec:limitations}: the soft-vote panel composition and the four-vs-five-seed polarity cap were both selected on dev, whereas detection's only dev-selected hyperparameter is the cross-task stacker rule (the BERT detector itself trains for a fixed epoch budget with no early stopping). Within extraction, precision and recall are close, while the token-level IoU indicates that on average predicted and gold token sets overlap less than half of their union (Table~\ref{tab:test}); this aggregate mixes boundary misalignment with full span misses and cannot by itself separate the two. We do not compare the test token IoU to the dev overlap \Fone{} because the two are different relaxations of exact match. The gaps may also reflect domain shift on the unseen test split, but we have no evidence for a specific shift and do not claim one.

In the organizers' final evaluation across the two participating teams and the organizers' baseline~\cite{kiesel:2026d}, our system scores highest on extraction (granularity-adjusted \Fone{} 0.728, against 0.665 for the baseline and 0.587 for the other team) and highest among the teams on polarity (macro \Fone{} 0.817 against 0.778, with the baseline at 0.839). On detection it ranks third on \Fone{} (0.869 against 0.884 and 0.872), with the highest recall (0.890) and the lowest precision of the three systems.

\section{Discussion and Future Work}
\label{sec:discussion}

\paragraph{Statistical strength of the NLI gain.}
The post-hoc NLI verifier moves dev exact \Fone{} from 0.5933 to
0.5953 ($+0.0020$, at a $-0.0019$ overlap-\Fone{} cost;
Table~\ref{tab:st2-progression}), within noise, and we did not
run a paired bootstrap or McNemar test; we therefore treat the
gain as direction-consistent rather than statistically
certified. The qualitative evidence agrees: all three spans
dropped at $\tau=0.1$ are truncated fragments of gold spans,
false positives under the exact metric that still carried
partial overlap credit, which is why removing them helps exact
\Fone{} while slightly lowering overlap \Fone{}
(Section~\ref{sec:st2-nli}).

\paragraph{Future work.}
Two threads follow from the present results. The first targets
polarity supervision by adding a rule-based negation-scope feature
to the polarity model, augmenting the input with one extra
per-token label (``in negation scope?''); this is motivated by the
67.8\% counter-causal versus 15.8\% procausal negation-cue
prevalence gap documented in Section~\ref{sec:data}. The second
targets the extraction backbone and candidate generation.
Alternative encoders include ELECTRA-large, structurally distinct
from RoBERTa's MLM objective; XLM-RoBERTa-large~\cite{conneau:2020xlmr},
pretrained on CC-100 across 100 languages and potentially better
suited to Indian-English news register; and NER architectures
that decouple span-boundary modeling from interior tagging,
whether through a locate-then-label boundary-regression
stage~\cite{shen:2021ll}, word-pair relation
classification~\cite{li:2022w2ner}, or biaffine span
scoring~\cite{yu:2020biaffine}. A
separate direction restricts extraction-stage span candidates to
constituent edges from a parser, which would render many one-sided
clause-attachment errors structurally impossible when the parser
is correct.

\section{Limitations}
\label{sec:limitations}

We disclose the principal threats to validity rather than relegate them to a footnote. The body of Sections~\ref{sec:st1}--\ref{sec:st3} reports CCNC development numbers; the held-out TIRA test scores are lower on every subtask (Table~\ref{tab:test}), and the comparison against Hagen et al.\ assumes their exact-match definition matches ours, which we have not independently verified. Nearly every discrete design choice, from the soft-vote panel and the stacker rule to the NLI threshold, the augmentation dose, the input-marker format, and the data-inclusion decisions, was selected on the same development split with no held-out internal validation; the 12.0\pp{} dev-to-test gap on extraction, the largest of the three, is the most plausible aggregate signature of this selection.

Statistical support for the smaller gains is correspondingly thin. The post-hoc NLI verifier yields a within-noise exact-\Fone{} gain on dev, which we report as direction-consistent rather than certified (Section~\ref{sec:discussion}). Per-seed variance is available only for the polarity ensemble (mean 0.8836, sample standard deviation 0.0057 over four seeds); the soft-vote gain on extraction and the stacker gain on detection are single-run point estimates without variance bands.

CCNC, drawn from Indian-English news outlets, is the only corpus we evaluate on, so transfer of the soft-vote and augmentation findings to other domains, English varieties, or non-news genres is untested. Finally, although the augmentation prompts draw exclusively on the CCNC training split, so held-out sentences cannot enter the training data by construction, GPT-5.4 may have seen CCNC or Hagen et al.'s pattern exemplars during pretraining; we did not run overlap checks of the generated outputs against the dev or test splits, and the structural validators certify tag preservation, not semantic counter-causality.

\section{Conclusion}
\label{sec:conclusion}

We described a system for the Touch\'e 2026 Causality Extraction task on the Countercausal News Corpus. On the held-out test set it reaches binary \Fone{} 0.869 on detection and macro \Fone{} 0.817 on polarity, and in the organizers' final causal-only evaluation of extraction it scores granularity-adjusted \Fone{} 0.728, the highest extraction score among all submissions including the organizers' baseline (Section~\ref{sec:test}). The development split, used for component selection and ablations, scores higher throughout (Table~\ref{tab:test}), including exact-match extraction \Fone{} 0.5933 against 0.473 under TIRA's all-samples exact-match metric on test, which a post-hoc DeBERTa-MNLI span verifier raises further within the noise band on dev (Section~\ref{sec:discussion}).

The system combines a three-layer BILOU+CRF stack on RoBERTa-large with emission-level soft-vote ensembling across three seeds, a single-rule cross-task stacker in which extraction-stage spans gate detection-stage false positives, and a nine-pattern counter-causal augmentation that adds 2{,}433 polarity training examples. The contribution is empirical rather than architectural: to our knowledge, neither emission-level CRF soft-voting with matched-panel measurement on the Causal News Corpus family nor a Hagen-style counter-causal pattern taxonomy applied to LLM-based augmentation on CCNC has been reported before. The single-rule cross-task stacker is the cross-task interaction we have the most evidence on: it added 2.6\pp{} on dev detection causal-class \Fone{}, but the dev-to-test drop on detection is of the same magnitude, so we cannot attribute net transfer to the rule without an unstacked test baseline. We retain it because precision-recall asymmetry favors precision and because the stacker did not degrade performance on any split we tested.

\begin{acknowledgments}
We thank the Touch\'e organizers for the lab and the evaluation infrastructure~\cite{froebe:2023}, and Hagen et al.~\cite{hagen:2025} for releasing both the CCNC corpus and a public baseline. Compute was provided by personal hardware and by University of Zurich shared GPU resources.
\end{acknowledgments}

\section*{Declaration on Generative AI}
During the preparation of this work, the authors used OpenAI GPT-5.4 for data generation, producing the nine-pattern counter-causal training examples for the polarity task described in Section~\ref{sec:st3-aug} under the structural validation reported there, and Claude (Anthropic) for code-generation assistance on training and evaluation scripts, which were human-reviewed and tested, and for writing assistance (grammar and spelling check, paraphrase and reword). All scientific decisions, analyses, and conclusions are the authors' own. After using these tools, the authors reviewed and edited the content as needed and take full responsibility for the publication's content.

\bibliography{references}

\appendix

\section{Training Hyperparameters}
\label{app:hyper}

Table~\ref{tab:hyper} lists the fine-tuning configuration of the
three deployed models. The TAPT encoder used by extraction seed
42 is produced beforehand by 50 epochs of masked-language
modeling on the 4{,}327-sentence in-domain corpus (learning rate
$10^{-4}$, 15\% token masking). Full configuration files ship
with the released repository.

\begin{table}[htbp]
  \caption{Fine-tuning hyperparameters per subtask model. Cells
  marked --- do not apply; unlisted settings follow library
  defaults.}
  \label{tab:hyper}
  \centering
  \begin{tabular}{lccc}
    \toprule
    Parameter & Detection & Extraction (per seed) & Polarity (per seed) \\
    \midrule
    Encoder & BERT-base & RoBERTa-large & BERT-base \\
    Optimizer & AdamW & AdamW ($\varepsilon=10^{-6}$) & AdamW \\
    Learning rate & $2\!\times\!10^{-5}$ & $5\!\times\!10^{-5}$ & $2\!\times\!10^{-5}$ \\
    CRF-group learning rate & --- & $3\!\times\!10^{-4}$ & --- \\
    Weight decay & 0.01 & 0.01 & --- \\
    Schedule & 10\% linear warmup, cosine & 6\% warmup, cosine & --- \\
    Gradient clip & 1.0 & 1.0 & --- \\
    Batch size & 32 & 16 & 32 \\
    Max sequence length & 128 & 256 & 256 \\
    Dropout & model default & 0.1 & model default \\
    Epochs & 5 (fixed) & $\leq$30 & $\leq$12 \\
    Early stopping & none & patience 5, overlap \Fone{} & patience 4, macro \Fone{} \\
    Seeds & 42 & 123, 2026, 42 & 42, 123, 2026, 7 \\
    EDA ($\alpha_{\text{sr}}$ / $\alpha_{\text{ri}}$) & --- & 0.4 / 0.5 & --- \\
    Aux.\ label smoothing & --- & $\varepsilon=0.1$, weight 0.2 (seed 42) & --- \\
    \bottomrule
  \end{tabular}
\end{table}

\section{Reproducibility Details}
\label{sec:reproducibility}

\paragraph{Compute and release.}
All models were fine-tuned on a single NVIDIA RTX 5090 (32~GB);
the seven deployed seeds complete in under twelve GPU-hours, and
full three-subtask inference on the development split runs in
under sixty seconds. The stack is PyTorch 2.4.1 (CUDA 12.1),
transformers 4.46.3, tokenizers 0.20.3, and pytorch-crf 0.7.2,
with full version pins in the released repository at
\url{https://github.com/rzninvo/NegaCausal}, which also hosts
the augmentation prompts and generated counter-causal data, the
inference and soft-vote code, the TIRA Docker recipe, the
per-seed checkpoints, and the stacker calibration; it will be
tagged at the CEUR camera-ready version. CCNC is distributed by
the Touch\'e organizers under the lab data-use agreement and
used only for CLEF 2026 participation;
CNCv2/RECESS~\cite{tan:2023recess} and
UniCausal~\cite{tan:2023unicausal} are used train-split-only
under their upstream licenses, the DeBERTa-v3 NLI
checkpoint~\cite{laurer:2023deberta} under its MIT license, and
we redistribute none of these.

\paragraph{Counter-causal data generation.}
The 2{,}433 augmented examples were generated by \mbox{OpenAI's}
GPT-5.4 via the Chat Completions API (JSON-response mode,
temperature 0.9, top\_p 1.0, max\_completion\_tokens 300); the
snapshot identifier and per-pattern prompt templates are in the
repository. Seeds and the single fixed exemplar come exclusively
from the CCNC training split, so no dev or test sentence appears
in any prompt. Each one-shot prompt rewrites an in-domain seed
into the target pattern while preserving the \texttt{<e0>} and
\texttt{<e1>} span contents verbatim; candidates are validated
for byte-equal tag preservation and exactly one cause/effect
pair (93.2\% acceptance, 2{,}433 of 2{,}610 attempts, a
syntactic rather than semantic pass rate).

\end{document}